\pdfoutput=1

\documentclass[11pt]{article}

\usepackage{EMNLP2022}

\usepackage{times}
\usepackage{latexsym}

\usepackage[T1]{fontenc}
\usepackage[utf8]{inputenc}
\usepackage{microtype}
\usepackage{multirow}
\usepackage{arydshln}
\usepackage{subcaption}
\usepackage{caption}
\usepackage{graphicx}
\usepackage{adjustbox}
\usepackage{float}
\usepackage{amsmath}
\usepackage{balance} 
\usepackage{amssymb}
\usepackage{inconsolata}
\title{Towards Multi-Modal Sarcasm Detection via Hierarchical Congruity Modeling with Knowledge Enhancement}

 \author{Hui Liu\textsuperscript{1} \quad \quad  \quad \quad 
  Wenya Wang\textsuperscript{2,3} \quad \quad \quad \quad  Haoliang Li\textsuperscript{1}\\
  \textsuperscript{1}City University of Hong Kong \\
    \textsuperscript{2}Nanyang Technological University\\
  \textsuperscript{3}University of Washington \\ 
\text{liuhui3-c@my.cityu.edu.hk}, \text{wangwy@ntu.edu.sg}, \text{haoliang.li@cityu.edu.hk}
}
\begin{document}
  \maketitle
\begin{abstract}
Sarcasm is a linguistic phenomenon indicating a discrepancy between literal meanings and implied intentions. Due to its sophisticated nature, it is usually challenging to be detected from the text itself. As a result, multi-modal sarcasm detection has received more attention in both academia and industries. However, most existing techniques only modeled the atomic-level inconsistencies between the text input and its accompanying image, ignoring more complex compositions for both modalities. Moreover, they neglected the rich information contained in external knowledge, e.g., image captions. In this paper, we propose a novel hierarchical framework for sarcasm detection by exploring both the atomic-level congruity based on multi-head cross attention mechanism and the composition-level congruity based on graph neural networks, where a post with low congruity can be identified as sarcasm. In addition, we exploit the effect of various knowledge resources for sarcasm detection. Evaluation results on a public multi-modal sarcasm detection dataset based on Twitter demonstrate the superiority of our proposed model. 
\end{abstract}

\section{Introduction}

Sarcasm refers to satire or ironical statements where the literal meaning of words is contrary to the authentic intention of the speaker to insult someone or humorously criticize something. Sarcasm detection has received considerable critical attention because sarcasm utterances are ubiquitous in today's social media platforms like Twitter and Reddit.  However, it is a challenging task to distinguish sarcastic posts to date in light of their highly figurative nature and intricate linguistic synonymy \cite{multi_modality_intro1, text_modality_intro1}.

\begin{figure}[t]
\includegraphics[width=\columnwidth]{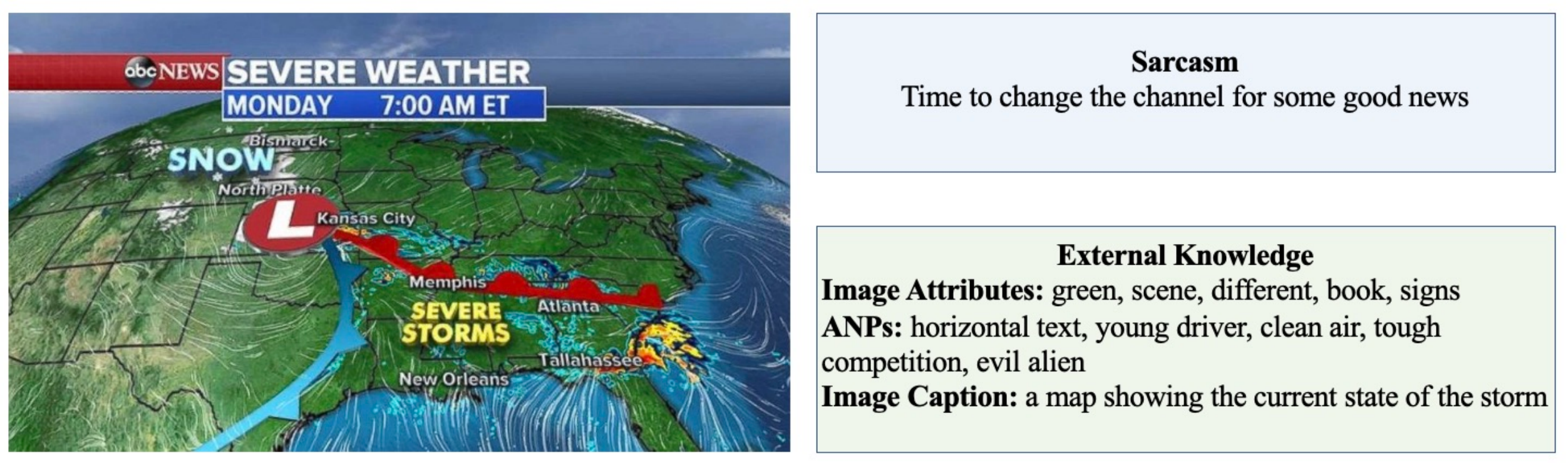}
\centering
\caption{An example of sarcasm along with the corresponding image and different types of external knowledge extracted from the image. The sarcasm sentence represents the need for some good news. However, the image of the TV program is switched to bad news depicting severe storms (bad weather) which contradicts the sentence.}
\label{sarcasm_image}
\vspace{-10pt}
\end{figure}

Early sarcasm detection methods mainly relied on fixed textual patterns, e.g., lexical indicators, syntactic rules, specific hashtag labels and emoji occurrences \cite{syntactic, hashtag, emoji}, which usually had poor performances and generalization abilities by failing to exploit contextual information. To resolve this issue, \cite{text_modality_intro1,text_modality_intro_2,text_modality_intro_3,text_modality_intro4} considered sarcasm contexts or the sentiments of sarcasm makers as useful clues to model congruity level within texts to gain consistent improvement. However, purely text-modality-based sarcasm detection methods may fail to discriminate certain sarcastic utterances as shown in Figure \ref{sarcasm_image}. In this case, it is hard to identify the actual sentiment of the text in the absence of the image forecasting severe weather. As text-image pairs are commonly observed in the current social platform, multi-modal methods become more effective for sarcasm prediction by capturing congruity information between textual and visual modalities \cite{multi_modality_intro1, multi_modality_mao, multi_modality_org, multi_modality_quantum, multi_modality_state_of_art, multi_modality_twitter_dataset}. 

However,  most of the existing multi-modal techniques only considered the congruity level between each token and image-patch \cite{multi_modality_mao,text_modality_intro1} and ignored the importance of multi-granularity (e.g., granularity such as objects, and relations between objects) alignments, which have been proved to be effective in other related tasks, such as cross-modal retrieval \cite{retrieval} and image-sentence matching \cite{matching, matching2}. In fact, the hierarchical structures of both texts and images advocate for composition-level modeling besides single tokens or image patches \cite{hier1}. By exploring compositional semantics for sarcasm detection, it helps to identify more complex inconsistencies, e.g., inconsistency between a pair of related entities and a group of image patches. 

Moreover, as figurativeness and subtlety inherent in sarcasm utterances may bring a negative impact to sarcasm detection, some works \cite{knowledge_text,knowledge_text1} found that the identification of sarcasm also relies on the external knowledge of the world beyond the input texts and images as new contextual information. What's more, it has drawn increasing research interest in how to incorporate knowledge to boost many machine learning algorithms such as recommendation system \cite{sun2021personalized} and 
 relation extraction \cite{sun2022conlearn}. Indeed, several studies extracted image attributes \cite{multi_modality_twitter_dataset} or adjective-noun pairs (ANPs) \cite{multi_modality_mao} from images as visual semantic information to bridge the gap between texts and images. However, constrained by limited training data, such external knowledge may not be sufficient or accurate to represent the images (as shown in Figure \ref{sarcasm_image}) which may bring negative effects for sarcasm detection. Therefore, how to choose and leverage external knowledge for sarcasm detection is also worth being investigated. 

To tackle the limitations mentioned above, in this work, we propose a novel hierarchical framework for sarcasm detection. Specifically, our proposed method takes both atomic-level congruity between independent image objects and tokens, as well as composition-level congruity considering object relations and semantic dependencies to promote multi-modal sarcasm identification. To obtain atomic-level congruity, we first adopt the multi-head cross attention mechanism \cite{attention} to project features from different modalities into the same space and then compute a similarity score for each token-object pair via inner products. Next, we obtain composition-level congruity based on the output features of both textual modality and visual modality acquired in the previous step. Concretely, we construct textual graphs and visual graphs using semantic dependencies among words and spatial dependencies among regions of objects, respectively, to capture composition-level feature for each modality using graph attention networks \cite{gat}. Our model concatenates both atomic-level and composition-level congruity features where semantic mismatches between the texts and images in different levels are jointly considered. Specially, we elaborate the terminology used in our paper again: congruity represents the semantic consistency between image and text. If the meaning of the image and text pair is contradictory, this pair will get less congruity. Atomic is between token and image patch, and compositional is between a group of tokens (phrase) and a group of patches (visual object).

Last but not the least, we propose to adopt the pre-trained transferable foundation models (e.g., CLIP \cite{CLIP, GPT-2}) to extract text information from the visual modality as external knowledge to assist sarcasm detection. The rationality of applying transferable foundation models is due to their effectiveness on a comprehensive set of tasks (e.g.,  descriptive and objective caption generation task) based on the zero-shot setting. As such, the extracted text contains ample information of the image which can be used to construct additional discriminative features for sarcasm detection. Similar to the original textual input, the generated external knowledge also contains hierarchical information for sarcasm detection which can be consistently incorporated into our proposed framework to compute multi-granularity congruity against the original text input.

The main contributions of this paper are summarized as follows: 1) To the best of our knowledge, we are the first to exploit hierarchical semantic interactions between textual and visual modalities to jointly model the atomic-level and composition-level congruities for sarcasm detection; 2) We propose a novel kind of external knowledge for sarcasm detection by using the pre-trained foundation model to generate image captions which can be naturally adopted as the input of our proposed framework; 3) We conduct extensive experiments on a publicly available multi-modal sarcasm detection benchmark dataset showing the superiority of our method over state-of-the-art methods with additional improvement using external knowledge.

\section{Related Work}


\subsection{Multi-modality Sarcasm Detection} 

With the rapid growth of multi-modality posts on modern social media, detecting sarcasm for text and image modalities has increased research attention.  \citet{multi_modality_org} first defined multi-modal sarcasm detection task. \citet{multi_modality_twitter_dataset} created a multi-modal sarcasm detection dataset based on Twitter and proposed a powerful baseline fusing features extracted from both modalities. \citet{multi_modality_mao} modeled both cross-modality contrast and semantic associations by constructing the Decomposition and Relation Network to capture commonalities and discrepancies between images and texts. 
\citet{multi_modality_intro1} and \citet{multi_modality_state_of_art} modeled intra-modality and inter-modality incongruities utilizing transformers \cite{attention} and graph neural networks, respectively. However, these works neglect the important associations played by hierarchical or multi-level cross-modality dismatches. To address this limitation, we propose to capture multi-level associations between modalities by cross attentions and graph neural networks to identify sarcasm in this work. 

\subsection{Knowledge Enhanced Sarcasm Detection}

\citet{knowledge_text} and \citet{knowledge_text1} pointed out that commonsense knowledge is crucial for sarcasm detection. For multi-modal based sarcasm detection,  \citet{multi_modality_twitter_dataset} proposed to predict five attributes for each image based on the pre-trained ResNet model \cite{resnet} as the third modality for sarcasm detection. In a similar fashion, \citet{multi_modality_mao} extracted adjective-noun pairs(ANPs) from every image to reason discrepancies between texts and ANPs. In addition, as some samples can contain text information for the images, \citet{multi_modality_intro1} and \citet{ multi_modality_state_of_art} proposed to apply the Optical Character Recognition (OCR) to acquire texts on the images. More recently, \citet{liang-etal-2022-multi} proposed to incorporate objection detection framework and label information of detected visual objects to mitigate modality gap. However, the knowledge extracted from these methods is either not expressive enough to convey the information of the images or is only restricted to a fixed set, e.g., nearly one thousand classes for image attributes or ANPs. Moreover, it should be noted that not every sarcasm post has text on images. To this end, in this paper, we propose to generate a descriptive caption with rich semantic information for each image based on the pre-trained Clipcap model \cite{clipcap}, which uses the CLIP \cite{CLIP} encoding as a prefix to the caption by employing a simple mapping network and then fine-tunes GPT-2 \cite{GPT-2} to generate the image captions.

\section{Methodology}

Our proposed framework contains four main components: Feature Extraction, Atomic-Level Cross-Modal Congruity, Composition-Level Cross-Modal Congruity and Knowledge Enhancement. Given an input text-image pair, the feature extraction module aims to generate text features and image features via a pre-trained text encoder and an image encoder, respectively. These features will then be fed as input to the atomic-level cross-modal congruity module to obtain congruity scores via a multi-head cross attention model (MCA). To produce composition-level congruity scores, we construct a textual graph and a visual graph and adopt graph attention networks (GAT) to exploit complex compositions of different tokens as well as image objects. The input features to the GAT are taken from the output of the atomic-level module. Due to the page limitation, we place our illustration figure in Figure \ref{overview}. Our model is flexible to incorporate external knowledge as a "virtual" modality, which could be used to generate complementary features analogous to the image modality for congruity score computation.

\subsection{Task Definition \& Motivation}
Multi-modal sarcasm detection aims to identify whether a given text associated with an image has a sarcastic meaning. Formally, given a multi-modal text-image pair $(X_T, X_I)$, where $X_T$ corresponds to a textual tweet and $X_I$ is the corresponding image, the goal is to produce an output label $y\in\{0,1\}$, where $1$ indicates a sarcastic tweet and $0$ otherwise. The goal of our model is to learn a hierarchical multi-modal sarcasm detection model (by taking both atomic-level and composition-level congruity into consideration) based on the input of textual modality, image modality and the external knowledge if chosen.  

\begin{figure}[t]
  \centering
  \includegraphics[width=\columnwidth]{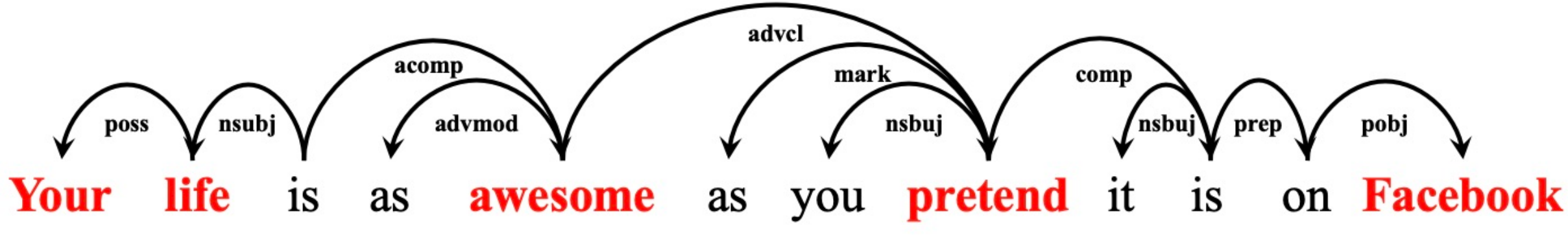}
  \caption{Semantic dependency between words in sarcasm text}
\label{dependency}
\vspace{-10pt}
\end{figure}

The reason to use composition-level modeling is to cope with the complex structures inherent in two modalities. For example, as shown in Figure \ref{dependency}, the semantic meaning of the sentence depends on composing \textit{your life}, \textit{awesome} and \textit{pretend} to reflect a negative position, which could be reflected via the dependency graph. The composed representation for text could then be compared with the image modality for more accurate alignment detection.

\subsection{Feature Extraction}
Given an input text-image pair $(X_T,X_I)$, where $X_T = \{w_1, w_2, \ldots, w_n\}$ consists of $n$ tokens, we utilize the pre-trained BERT model \cite{bert} with an additional multi-layer perceptron (MLP) to produce a feature representation for each token, denoted as $\mathbf{T}=[\mathbf{t}_1,  \mathbf{t}_2, \ldots, \mathbf{t}_{n}]$, 
where $\mathbf{T} \in \mathbb{R}^{n\times d}$. As for image processing, given the image $X_I$ with the size ${L_h\times L_w}$, following existing methods \cite{multi_modality_mao,multi_modality_twitter_dataset,multi_modality_state_of_art,multi_modality_intro1}, we first resize the image to size $224\times224$. Then we divide each image into $r$ patches and reshape these patches into a sequence, denoted as $\left\{p_1, p_2, \ldots, p_r \right\}$, in the same way as tokens in the text domain. Next, we feed the sequence of $r$ image patches into an image encoder to get a visual representation for each patch. Specifically, in this paper, we choose two kinds of image encoders including the pre-trained Vision Transformer (ViT) \cite{ViT} and a ResNet model \cite{resnet}, both of which are trained for image classification on ImageNet. Hence, the embedding of image patches derived by ViT or ResNet contains rich image label information. Here we adopt the features before the final classification layer to initialize the embeddings for visual modality. We further use a two-layer MLP to obtain the feature representations for $\left\{p_1, p_2, \ldots, p_r \right\}$ as  $\mathbf{I} = [\mathbf{i}_1, \mathbf{i}_2,\ldots, \mathbf{i}_{r}]$, where $\mathbf{I} \in \mathbb{R}^{r\times d}$. 

\subsection{Atomic-Level Congruity Modeling}
To measure atomic-level congruity between a text sequence and an image, an intuitive solution is to compute a similarity score between each token and a visual patch directly. However, due to the huge gap between two different modalities, we propose to use cross attention mechanisms with $h$ heads to firstly align the two modalities in the same space, which can be computed as
\begin{equation}\label{eq:head}
\small
\textbf{\textrm{head}}_i =\textrm{softmax}\left (\frac{(\mathbf{T} \mathbf{W}_q^i)^\top} {\sqrt{{d}/{h}}} (\mathbf{I} \mathbf{W}_k^i)\right ) (\mathbf{I}  \mathbf{W}^i_v),
\end{equation}
where $\mathbf{I}\in \mathbb{R}^{r\times d}$ and $ \mathbf{T} \in\mathbb{R}^{n\times d}$ are feature representations of the given text and image, respectively.  $\mathbf{W}_q^i\in \mathbb{R}^{d\times\frac{d}{h}}$, $\mathbf{W}_k^i\in \mathbb{R}^{d\times\frac{d}{h}}$ and $\mathbf{W}_v^i\in \mathbb{R}^{d\times\frac{d}{h}}$ are query, key and value projection matrices, respectively, for $\textbf{\textrm{head}}_i\in \mathbb{R}^{n\times\frac{d}{h}}$. It is worth noting that we also consider taking image as query, text as key and value for Equation (\ref{eq:head}). However, we empirically find that the performance is not desired in this case. We conjecture the reason to be the fact that the visual modality may not contain sufficient information and is less expressive compared to the textual modality to provide attentive guidance, which can lead to negative impact of the final performance.

Then, by concatenating all heads followed by a two-layer MLP and a residual connection, we obtain updated text representations $\tilde{\mathbf{T}} \in \mathbb{R}^{n\times d}$ after aligning with the visual modality as 
\begin{equation}\label{eq:norm}
\small
\tilde{\mathbf{T}} = \textrm{norm}(\mathbf{T}+\textrm{MLP}([\textbf{\textrm{head}}_1\, \Vert \, \textbf{\textrm{head}}_2 \, \Vert \,\ldots \, \Vert \, \textbf{\textrm{head}}_h])),
\end{equation}
where ``$\textrm{norm}$" denotes the layer normalization operation and ``$\Vert$" denotes the concatenation operation. Next, to perform atomic-level cross-modal congruity detection, we adopt the inner product as $\mathbf{Q}_a = \frac{1}{\sqrt{d}}(\tilde{\mathbf{T}}{\mathbf{I}}^{\top})$
where $\mathbf{Q}_a \in \mathbb{R}^{n\times r}$ is the matrix consisting of ${\mathbf{Q}_a[i,j]}$ for $i$-th row and $j$-th column representing the similarity score between the $i$-th token of the text and the $j$-th patch of the image. Intuitively, different words can have different influence on the sarcasm detection task. For example, noun, verb and adjacent words are usually more important for understanding sarcastic utterances. As such, we feed features of words to a fully-connected (FC) layer with a softmax activation function to model the token importance for sarcasm detection. The final atomic level congruity score $\mathbf{s}_a$ can be obtained by a weighted sum of $\mathbf{Q}_a$ with the importance score of each token as
\begin{equation}
\small
\mathbf{s}_a = \textrm{softmax}(\tilde{\mathbf{T}} \mathbf{W}_a + \mathbf{b}_a)^\top \mathbf{Q}_a, \label{last token incongruity}
\end{equation}
where $\mathbf{W}_a \in \mathbb{R}^{d\times 1}$ and $\mathbf{b}_a \in \mathbb{R}^n$ are trainable parameters in the FC layer for token importance score computation. $\mathbf{s}_a \in \mathbb{R}^r$ contains the predicted atomic-level congruity score corresponding to each of the $r$ patches.

\subsection{Composition-Level Congruity Modeling}

The composition-level congruity detection considers the more complex structure of both the text and image modalities, compared to the atomic-level computations. To achieve that, we propose to first construct a corresponding textual graph and a visual graph for the input text-image pair. For the textual graph, we consider tokens in the input text as graph nodes and use dependency relations between words extracted by spaCy\footnote{\url{https://spacy.io/}} as edges, which have been proved to be effective for various graph-related tasks \cite{matching2, multi_modality_state_of_art}. Concretely, if there exists a dependency relation between two words, there will be an edge between them in the textual graph. For the visual graph, given $r$ image patches, we take each patch as a graph node and connect adjacent nodes according to their geometrical adjacency. Additionally, both two kinds of graphs are undirected and contain self-loops for expressiveness.

Then, we model the graphs in text and visual modalities with graph attention networks (GAT) \cite{gat}. GAT leverages self-attention layers to weigh the extent of information propagated from corresponding nodes. 
By using GAT, atomic-level semantic information will propagate along with the graph edge to learn composition-level representations for both textual modality and image modality. Here, we take the textual graph for illustration given as
\begin{equation}
\small
\alpha^l_{i,j} =
\frac{
\exp\left(\textrm{LeakyReLU}\left(\mathbf{v}_l^\top
[\mathbf{\Theta}_l \mathbf{t}^l_i \Vert \, \mathbf{\Theta}_l \mathbf{t}^l_j]\right)\right)}
{\sum_{k}
\exp\left(\textrm{LeakyReLU}\left(\mathbf{v}_l^\top
[\mathbf{\Theta}_l \mathbf{t}^l_i \Vert \, \mathbf{\Theta}_l \mathbf{t}^l_k
]\right)\right)},
\end{equation}
\begin{equation}
\small
\mathbf{t}^{l+1}_i = \alpha^l_{i,i}\mathbf{\Theta}_l \mathbf{t}^l_i+\sum_{j \in \mathcal{N}(i)}\alpha_{i,j}\mathbf{\Theta}_l \mathbf{t}^l_{j},
\end{equation}
where $k \in \mathcal{N}(i) \cup \{ i \}$, $\mathbf{\Theta}_l  \in \mathbb{R}^{d \times d}$ and $\mathbf{v}_l \in \mathbb{R}^{2d}$ are learnable parameters of the $l$-th textual GAT layer. $\alpha^l_{i,j}$ is a scalar indicating the attention score between node $i$ and its neighborhood node $j$. $\mathbf{t}^l_i$ represents the feature of node $i$ in the $l$-th layer, with $\mathbf{t}^0_i=\tilde{\mathbf{t}}_i$ initialized from the atomic-level features $\tilde{\mathbf{T}}$. 
We use $\hat{\mathbf{T}}= [\mathbf{t}^{L_T}_1, \mathbf{t}^{L_T}_2, \ldots, \mathbf{t}^{L_T}_r]$ with $\hat{\mathbf{T}} \in \mathbb{R}^{n \times d}$ to represent the composition-level embeddings of the textual modality after $L_T$ GAT layers that incorporate complex dependencies among related tokens. In some cases, we may not be able to construct a reliable textual graph  due to the lack of sufficient words in a sentence or errors from the parser. Hence, we further propose to concatenate $\hat{\mathbf{T}}$ with a sentence embedding  $\mathbf{c} \in \mathbb{R}^d$ which is computed by a weighted sum of each word embedding in $\tilde{\mathbf{T}}$:
\begin{equation}
\label{eq666}
\small
    \mathbf{c} = \mathrm{softmax}(\mathbf{T} \mathbf{W}_c + \mathbf{b}_c)^\top \tilde{\mathbf{T}},
\end{equation} 
with learnable $\mathbf{W}_c \in \mathbb{R}^{d\times 1}$ and $\mathbf{b}_c \in \mathbb{R}^n$.



Likewise, we can obtain $\hat{\mathbf{I}}=[\mathbf{i}^{L_I}_1, \mathbf{i}^{L_I}_2,\ldots, \mathbf{i}^{L_I}_n]$, $\hat{\mathbf{I}} \in \mathbb{R}^{r \times d}$ as the composition-level representations in the visual modality. At last, we compute composition-level alignment scores $\mathbf{s} _p$ between $\hat{\mathbf{T}}$ and $\hat{ \mathbf{I} }$ in a similar way as atomic-level congruity as
\begin{equation}
\small
    \mathbf{s}_p = \textrm{softmax}([\hat{\mathbf{T}}\, \Vert \, \mathbf{c}] \mathbf{W}_p + \mathbf{b}_p)^\top \mathbf{Q}_p,
\end{equation}
where {\small $\mathbf{Q}_p = \frac{1}{\sqrt{d}}([\hat{ \mathbf{T} }\, \Vert \, \mathbf{c}] \hat{\mathbf{I}}^\top) \in\mathbb{R}^{(n+1)\times r}$}
is the matrix of composition-level congruity between textual modality and visual modality, $\mathbf{W}_p \in \mathbb{R}^{d\times 1}$ and $\mathbf{b}_p \in \mathbb{R}^{n+1}$ are trainable parameters. $\mathbf{s}_p \in \mathbb{R}^r$ contains the final predicted composition-level congruity score for each of the $r$ image patches.


\subsection{Knowledge Enhancement}
\label{knowledge_section}
While using text-image pair can benefit sarcasm detection  compared with only using a single modality
, recent works have shown that it might be still challenging to detect sarcasm solely from a text-image pair \cite{knowledge_text,  knowledge_text1}. 
To this end, we 
explore the effect of fusing
various external knowledge extracted from an image for sarcasm detection.
For example, the knowledge could be image attributes \cite{multi_modality_twitter_dataset}, ANPs \cite{multi_modality_mao} as they provide more information on the key concepts delivered by the image. However, such information lacks coherency and semantic integrity to describe an image and may introduce unexpected noise, as indicated in Figure \ref{sarcasm_image}. To address this limitation, we propose to generate image captions as the external knowledge to assist sarcasm detection. We further compare the effect of each knowledge form in the experiments.

To fuse external knowledge into our model, we treat knowledge $X_K$ as another ``virtual'' modality besides texts and images. Then the augmented input to the model becomes $(X_T,X_I, X_K)$. As the knowledge is given in textual form, we follow the process of generating text representations to attain the knowledge features. Specifically, we first obtain the input knowledge representations as $\mathbf{K} = [\mathbf{k}_1, \mathbf{k}_2, \ldots, \mathbf{k}_m]$ using BERT with a MLP, which is analogous to $\mathbf{T}$. Then, we propose to reason the congruity score between text and knowledge modalities at atomic-level by following the procedure of computing atomic-level congruity score between text and image modalities (as shown in Equations (\ref{eq:head})-(\ref{last token incongruity})) with another set of parameters. Concretely, for cross-modality attentions between texts and knowledge, we replace $\mathbf{I}$ in Equation (\ref{eq:head}) with $\mathbf{K}$ and $\mathbf{T}$ in Equation (\ref{eq:head}) with $\tilde{\mathbf{T}}$, which is the updated text representations after aligning with the visual modality. Inheriting information from the image modality, using $\tilde{\mathbf{T}}$ as the query to attend to knowledge enhances deeper interactions across all the three modalities. By further replacing $\mathbf{T}$ in Equation (\ref{eq:norm}) with $\tilde{\mathbf{T}}$, we denote the atomic-level text representations after aligning with the knowledge by $\tilde{\mathbf{T}}^k$. The similarity matrix between texts and knowledge becomes $\mathbf{Q}_a^k = \frac{1}{\sqrt{d}} (\tilde{\mathbf{T}}^k \mathbf{K}^{\top})$. 
Then the atomic-level congruity score, denoted as $\mathbf{s}^k_a \in \mathbb{R}^m$, can be obtained as 
\begin{equation}
\small
\mathbf{s}^k_a = \textrm{softmax}(\tilde{\mathbf{T}}^k \mathbf{W}^k_a + \mathbf{b}^k_a)^\top \mathbf{Q}^k_a. \label{knowledge token incongruity}
\end{equation}

By adopting the dependency graph for $X_K$, we further generate the updated knowledge representations $\hat{\mathbf{K}}$ via GAT and obtain the composition-level congruity score $\mathbf{s}^k_p \in \mathbb{R}^m$ between text and knowledge modalities, following the same procedure for text-image composition-level congruity score as described in Section 3.4. 
 

\subsection{Training \& Inference}
Given both the atomic-level and composition-level congruity scores $\mathbf{s}_a$ and $\mathbf{s}_p$, respectively, the final prediction could be produced considering the importance of each image patch for sarcasm detection.
{\small
\begin{eqnarray}
\mathbf{p}_v &=& \textrm{softmax}(\mathbf{I} \mathbf{W}_v + \mathbf{b}_v), \label{eq:p_v} \\
\mathbf{y}' &=& \textrm{softmax}(\mathbf{W}_y [\mathbf{p}_v \odot \mathbf{s}_a\, \Vert \, \mathbf{p}_v \odot \mathbf{s}_p] + \mathbf{b}_y), \label{eq:y}
\end{eqnarray}}where $\mathbf{W}_v \in \mathbb{R}^{d\times 1}$, $\mathbf{b}_v \in \mathbb{R}^r$, $\mathbf{W}_y \in\mathbb{R}^{2\times2r}$ and $\mathbf{b}_y \in \mathbb{R}^2$ are trainable parameters, $\mathbf{p}_v \in \mathbb{R}^r$ is a $r$-dim attention vector, $\odot$ is element-wise vector product. It is flexible to further incorporate external knowledge by reformulating Equation (\ref{eq:y}) to 

{\small
\begin{eqnarray}
    \mathbf{y}' = \textrm{softmax}(\mathbf{W}^k_y [\!\!\!\!\!\!\!\!\!\!\!\!&&\mathbf{p}_v \odot \mathbf{s}_a\, \Vert \, \mathbf{p}_v \odot \mathbf{s}_p\, \Vert \, \nonumber \\
    &&\mathbf{p}_k \odot \mathbf{s}^k_a\, \Vert \,\mathbf{p}_k \odot \mathbf{s}^k_p] + \mathbf{b}^k_v), \nonumber
\end{eqnarray}}
where $\mathbf{s}^k_a$ and  $\mathbf{s}^k_p$ are atomic-level and composition-level congruity scores between post and external knowledge. $\mathbf{p}_k \in \mathbb{R}^m$ measures the importance of each word in the knowledge obtained by {\small $\mathbf{p}_k = \textrm{softmax}(\mathbf{K} \mathbf{W}^k_v + \mathbf{b}^k_v)$}. The entire model can be trained in an end-to-end fashion by minimizing the cross-entropy loss given the ground-truth label $y$.



\section{Experiments}
\subsection{Dataset}

\begin{table}[H]
  \caption{Statistics of the dataset}
  \label{dataset}
  \begin{adjustbox}{max width=1.0\columnwidth}
  \begin{tabular}{c|ccc}
    \hline
    &Training&Development&Testing\\
    \hline
    Sarcasm&8642&959&959\\
    Non-Sarcasm&11174&1451&1450\\
    All&19816&2410&2409\\
    \hline
    Token Length&16.91&16.92&17.13\\
    Entity&3.76&3.71&3.84\\
    \hline
\end{tabular}
\end{adjustbox}
\end{table}

We evaluate our model on a publicly available multi-modal sarcasm detection dataset in English constructed by \citet{multi_modality_twitter_dataset}. The statistics are shown in Tabel \ref{dataset}. Based on our preliminary analysis, the average numbers of tokens and entities in a text are approximately $17$ and $4$, respectively, where complex compositions among atomic units have a higher chance of being involved. This finding provides the basis for our framework using atomic-level and composition-level information to capture hierarchical cross-modality semantic congruity.


\subsection{Implementation}
For a fair comparison, following the pre-processing in \cite{multi_modality_twitter_dataset, multi_modality_state_of_art, multi_modality_mao}, we remove samples containing words that frequently co-occur with sarcastic utterances (e.g., \textit{sarcasm}, \textit{sarcastic}, \textit{irony} and \textit{ironic}) to avoid introducing external information. The dependencies among tokens are extracted using spaCy toolkit. For image preprocessing, we resize the image to $224 \times 224$ and divide it into $32 \times 32$ patches (i.e., $p=7$, $r=49$). For knowledge extraction, we extract image attributes following \cite{multi_modality_twitter_dataset}, ANPs following \cite{multi_modality_mao} and image captions via Clipcap \cite{clipcap}.


Next, we employ a pre-trained BERT-base-uncased model\footnote{\url{https://huggingface.co/bert-base-uncased}} as textual backbone network to obtain initial embeddings for texts and knowledge, and choose the pre-trained  ResNet and ViT\footnote{\url{https://github.com/lukemelas/PyTorch-Pretrained-ViT}} modules as visual backbone networks to extract initial embeddings for images. These textual and visual representations are mapped to $200$-dim vectors by corresponding MLPs. We use Adam optimizer to train the model. The dropout and early-stopping are adopted to avoid overfitting. The details of implementations are listed in Table \ref{model parameters} in Appendix. Our code is avaliable at \url{https://github.com/less-and-less-bugs/HKEmodel}.


\subsection{Baseline Models}
We divide the baseline models into three categories: text-modality methods, image-modality methods and multi-modality methods. For text-based models, we adopt  \textbf{TextCNN} \cite{TextCNN}, \textbf{Bi-LSTM} \cite{bilstm}, 
\textbf{SMSD} \cite{text_modality_intro4} which adopts self-matching networks and low-rank bilinear pooling for sarcasm detection, and \textbf{BERT} \cite{bert} that generates predictions based on the [CLS] token as baseline models. For pure image-based models, we follow \cite{multi_modality_twitter_dataset, multi_modality_state_of_art} to utilize the feature representations after the pooling layer of \textbf{ResNet} and  [CLS] token in each image patch sequence obtained by \textbf{ViT} to generate predictions. For multi-modal based methods, we adopt \textbf{HFM} \cite{multi_modality_twitter_dataset}, \textbf{D\&R Net} \cite{multi_modality_mao}, \textbf{Att-BERT} \cite{multi_modality_intro1},  \textbf{InCrossMGs} \cite{multi_modality_state_of_art} and a variant of \textbf{CMGCN} \cite{liang-etal-2022-multi} without external knowledge as the multi-modal baselines.

\subsection{Results without External Knowledge}
\begin{table}
  \caption{Comparison results for sarcasm detection. $\dagger$ indicates ResNet backbone and $\ddagger$ indicates ViT backbone.}
  \vspace{-2mm}
  \centering
  \label{main results}
  \begin{adjustbox}{max width=1.0\columnwidth}
  \begin{tabular}{cc|cccc}
    \hline
    \multicolumn{2}{c}{Model}&Acc(\%)&P(\%)&R(\%)&F1(\%)\\
    \hline
    \multirow{4}*{Text}&TextCNN&80.03&74.29&76.39&75.32\\
    &Bi-LSTM&81.90&76.66&78.42&77.53\\
    &SMSD&80.90&76.46&75.18&75.82\\
    &BERT&83.85&78.72&82.27&80.22\\
    \hline
    \multirow{2}*{Image}&Image&64.76&54.41&70.80&61.53\\
    &ViT&67.83&57.93&70.07&63.43\\
    \hline
    \multirow{7}*{Multi-Modal}&$\textrm{HFM}^{\dagger}$&83.44&76.57&84.15&80.18\\
    &$\textrm{D\&R Net}^{\dagger}$&84.02&77.97&83.42&80.60\\
    &$\textrm{Att-BERT}^{\dagger}$&86.05&80.87&85.08&82.92\\
    &$\textrm{InCrossMGs}^{\ddagger}$&86.10&81.38&84.36&82.84\\
    &$\textrm{CMGCN}^{\ddagger}$&86.54&--&--&82.73\\
    \cdashline{2-6}[1pt/1pt]
    &$\textrm{Ours}^{\dagger}$&87.02&\textbf{82.97}&84.90&83.92\\
    &$\textrm{Ours}^{\ddagger}$&\textbf{87.36}&81.84&\textbf{86.48}&\textbf{84.09}\\
  \hline
\end{tabular}
\end{adjustbox}
\vspace{-10pt}
\end{table}

We first evaluate the effectiveness of our proposed framework by comparing with the baseline models as shown in Tabel \ref{main results}. It is shown that our proposed model achieves state-of-the-art performance. Obviously, text-based models perform far better than image-based methods, which implies that text is more comprehensible and more informative than images. This supports our intuition of extracting textual knowledge from images as additional clues. On the other hand, multi-modal methods outperform all those models in single modality. This illustrates that considering information from both modalities contributes to the task by providing additional cues on modality associations.

Note that compared with multimodal methods using ResNet as the visual backbone network, our model achieves a 0.97\% improvement in terms of Accuracy and a 1.00\% improvement in terms of F1-score over the state-of-art method \textbf{Att-BERT}. Besides, using ViT as the image feature extractor, our model outperforms the \textbf{InCrossMGs} model with a 1.26\% improvement in Accuracy and 1.25\% improvement in F1-score. Our method can also achieve better performance with improvement of 0.82\% based on Accuracy compared with the recent proposed \textbf{CMGCN}. The results demonstrate the effectiveness and superiority of our framework for sarcasm detection by modeling both atomic-level and composition-level cross-modality congruities in a hierarchical manner. 

\subsection{Results with External Knowledge}
\label{common_know}
\begin{table}
  \caption{Results of different knowledge types.}
  \vspace{-2mm}
    \centering
  \label{knowledge}
  \begin{adjustbox}{max width=0.8\columnwidth}
  \begin{tabular}{c|cc}
    \hline
    Knowledge Type&Acc(\%)&F1(\%)\\
    \hline
    w/o external knowledge &87.36&84.09\\
    \cdashline{1-3}
    Image Attributes&86.43&83.30\\
    ANPs &86.35&83.54\\
    Image Captions&\textbf{88.26}&\textbf{84.84}\\
    Image Captions (w/o image) &86.60&83.28\\
    \hline
\end{tabular}
\end{adjustbox}
\vspace{-10pt}
\end{table}

We then evaluate the effectiveness of our method by considering external knowledge. Table \ref{knowledge} reports the accuracy and F1-score for our proposed sarcasm detection method enhanced by considering different types of knowledge. By incorporating image captions, the performance further improves compared with the original model (w/o external knowledge). 
On the contrary, Image Attributes and ANPs bring negative effects and deteriorate the performance. We conjecture two possible reasons, 1) image attributes and ANPs can sometimes be meaningless or even noisy for identifying sarcasm, 2) image attributes and ANPs are rather short, lacking rich compositional information for our hierarchical model. Last but not the least, it is worth mentioning that only exploiting texts and captions in textual modality (Image Captions w/o image) without the visual modality also achieves superior performance compared with all multi-modal baselines in Table \ref{main results}. Such observation illustrates that the pre-trained models such as CLIP and GPT-2 can provide meaningful external information for sarcasm detection.


\subsection{Ablation Study}

\begin{table}
  \caption{Experimental results of ablation study.}
  \vspace{-2mm}
  \label{ablation}
    \centering
  \begin{adjustbox}{max width=0.75\columnwidth}
  \begin{tabular}{c|cc}
    \hline
    Model&Acc(\%)&F1(\%)\\
    \hline
    Ours&\textbf{87.36}&\textbf{84.09}\\
    \cdashline{1-3}
    w/o atomic-level&86.56&83.50\\
    w/o MCA and atomic-level&86.01&82.73\\
    w/o composition-level&82.60&79.13\\
    \hline
\end{tabular}
\end{adjustbox}
\vspace{-10pt}
\end{table}

\noindent\textbf{Impact of Different Components. }We conduct ablation studies using ViT as the visual backbone network without external knowledge to further understand the impact of different components of our proposed method. To be specific,  we consider three different scenarios, 1) remove the atomic-level congruity score $\mathbf{s}_a$ (denoted as w/o atomic-level), 2) 
remove both $\mathbf{s}_a$ and the multi-head cross attention (MCA) module by replacing $\tilde{T}$ to $T$ in all the computations (denoted as w/o MCA and atomic-level), 3) remove composition-level congruity score $\mathbf{s}_p$ (denoted as w/o composition-level).

The results are shown in Table \ref{ablation}. It is clear that our model achieves the best performance when composing all these components. 
It is worth noting that the removal of composition-level $\mathbf{s}_p$ leads to significant performance drop, compared to atomic-level removal. This indicates that the composition-level congruity plays a vital role for discovering inconsistencies between visual and textual modalities by exploiting complex structures through propagating atomic representations along the semantic or geographical dependencies. Moreover, the removal of MCA leads to slightly lower performance which indicates that cross attention is beneficial for modeling cross modality interactions and reducing the modality gap in the representation space.


\begin{figure}[t]
\centering
\begin{subfigure}{0.22\textwidth}
\centering
\includegraphics[width=\linewidth]{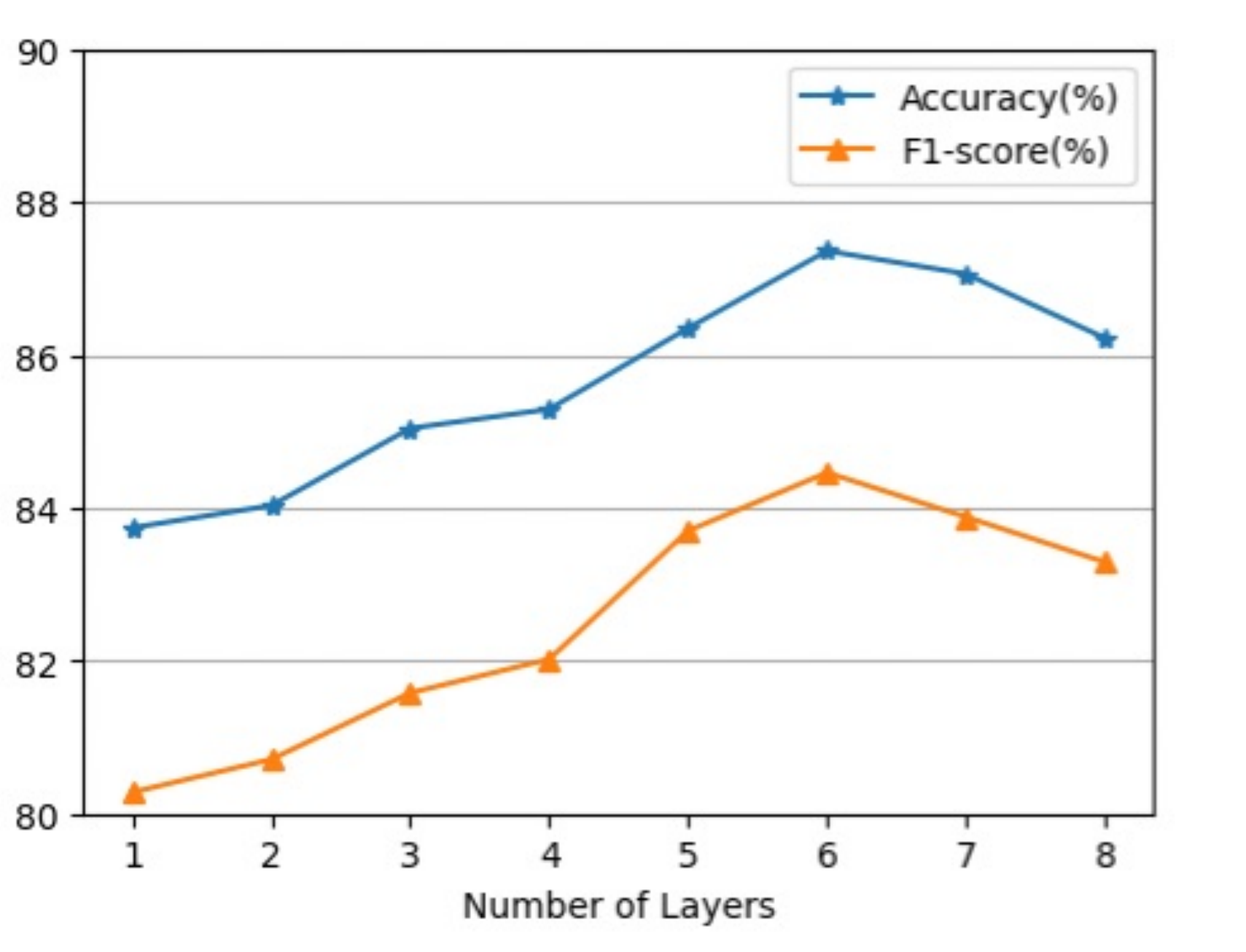}
\caption{MCA Layers}
\label{MCA}
\end{subfigure}
\begin{subfigure}{0.22\textwidth}
\includegraphics[width=\linewidth]{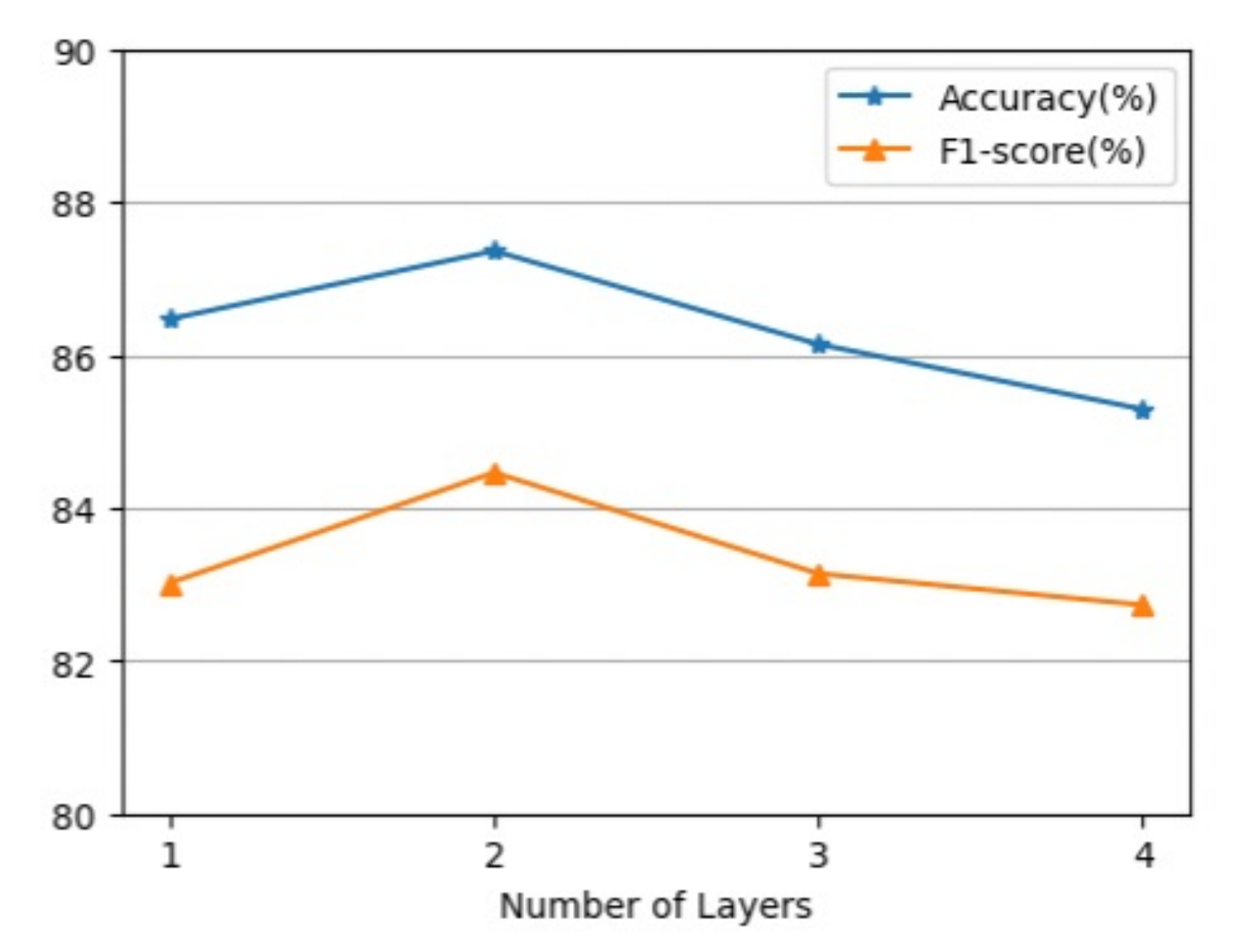} 
\centering
\caption{GAT Layers}
\label{gat}
\end{subfigure}
\caption{Performance of different model architectures.}
\label{architectures}
\vspace{-10pt}
\end{figure}
\noindent\textbf{Impact of MCA Layers.}
We measure the performance change without external knowledge when varying the number of MCA layers from 1 to 8 in Figure \ref{MCA}. As can be seen, the performance first increases along with the increasing number of layers and then decreases after 6 layers. This shows excessive MCA layers in atomic-level congruity module may overfit to textual and visual modality alignment instead of sarcasm detection.

\noindent\textbf{Impact of GAT Layers.}
We analyse the impact of the number of GAT layers for our proposed model and report Accuracies and F1 scores in Figure \ref{gat}. The results show that the best performance can be achieved when using a two-layer GAT model and the performance further drops when increasing the number of layers. We conjecture the reason to be the over-smoothing issue of Graph Neural Networks when increasing the number of proœpagation layers, making different nodes indistinguishable.

\noindent\textbf{Impact of Different Sentence Embeddings.}
\begin{table}
  \caption{Experimental results of different sentence embedding.}
  \label{ablation_sen}
    \centering
  \begin{adjustbox}{max width=1\columnwidth}
  \begin{tabular}{c|c|c|c|c}
     \hline
    Model&USE&Word Averaging&CLIP&Bert\\
    \hline
    Accuracy(\%)&86.98&87.02&\textbf{88.10}&87.10\\
    \hline
\end{tabular}
\end{adjustbox}
\vspace{-10pt}
\end{table}
We perform experiments using Universal Sentence Encoder (USE) \cite{use}, CLIP \cite{CLIP}, CLS Token of Bert \cite{bert}, and Word Averaging to extract sentence embedding (i.e., $\mathbf{c}$ in Eq \ref{eq666}) as shown in Tabel \ref{ablation_sen}. Although CLIP outperforms other methods by a slight margin, we prefer Word Averaging to keep our model concise and reduce the number of parameters.  
\subsection{Case Study}

\begin{figure}[t]
\centering
\begin{subfigure}{0.22\textwidth}
\centering
\includegraphics[width=\linewidth]{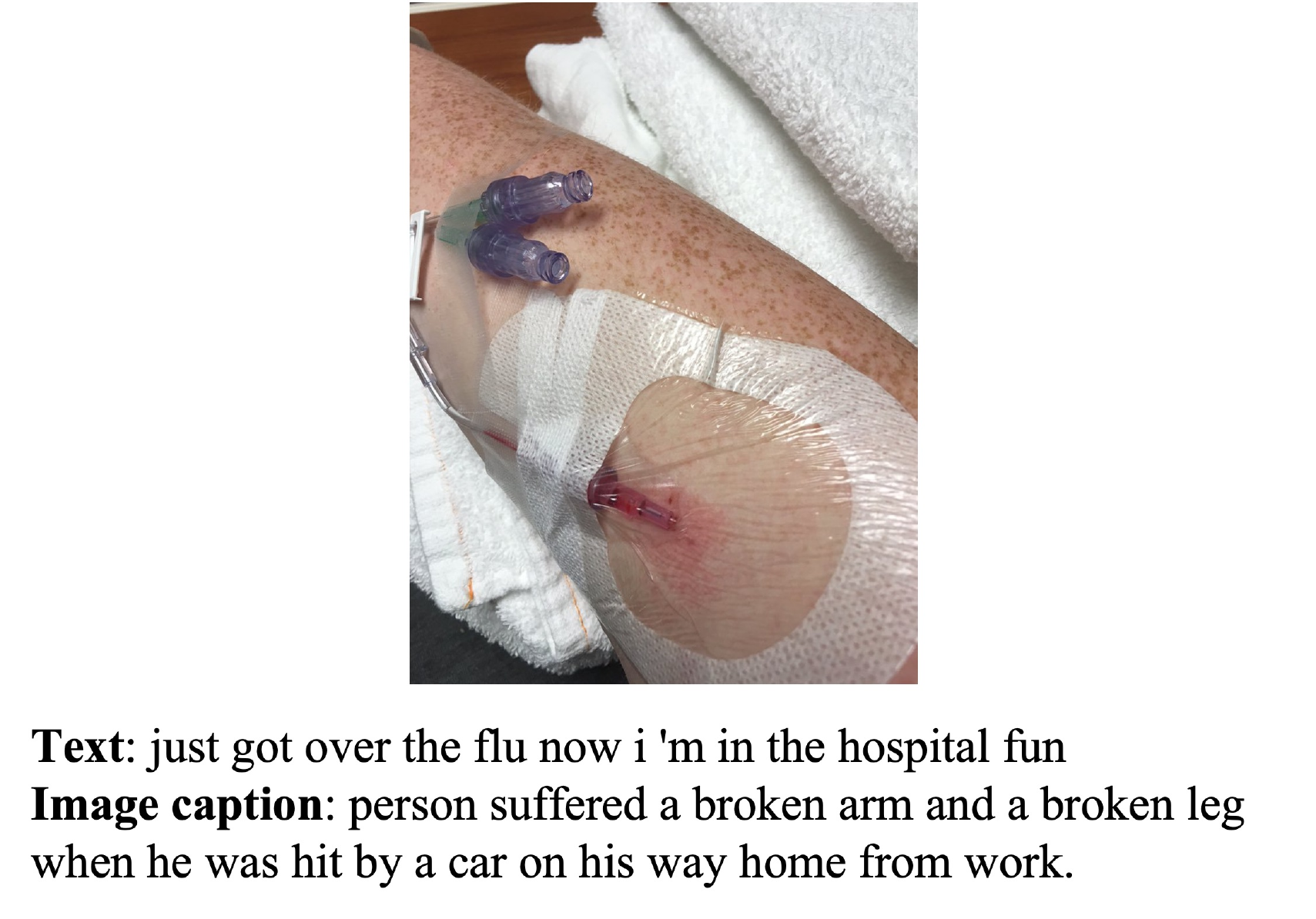}
\caption{Sarcasm}
\label{casestudy1}
\end{subfigure}
\begin{subfigure}{0.22\textwidth}
\includegraphics[width=\linewidth]{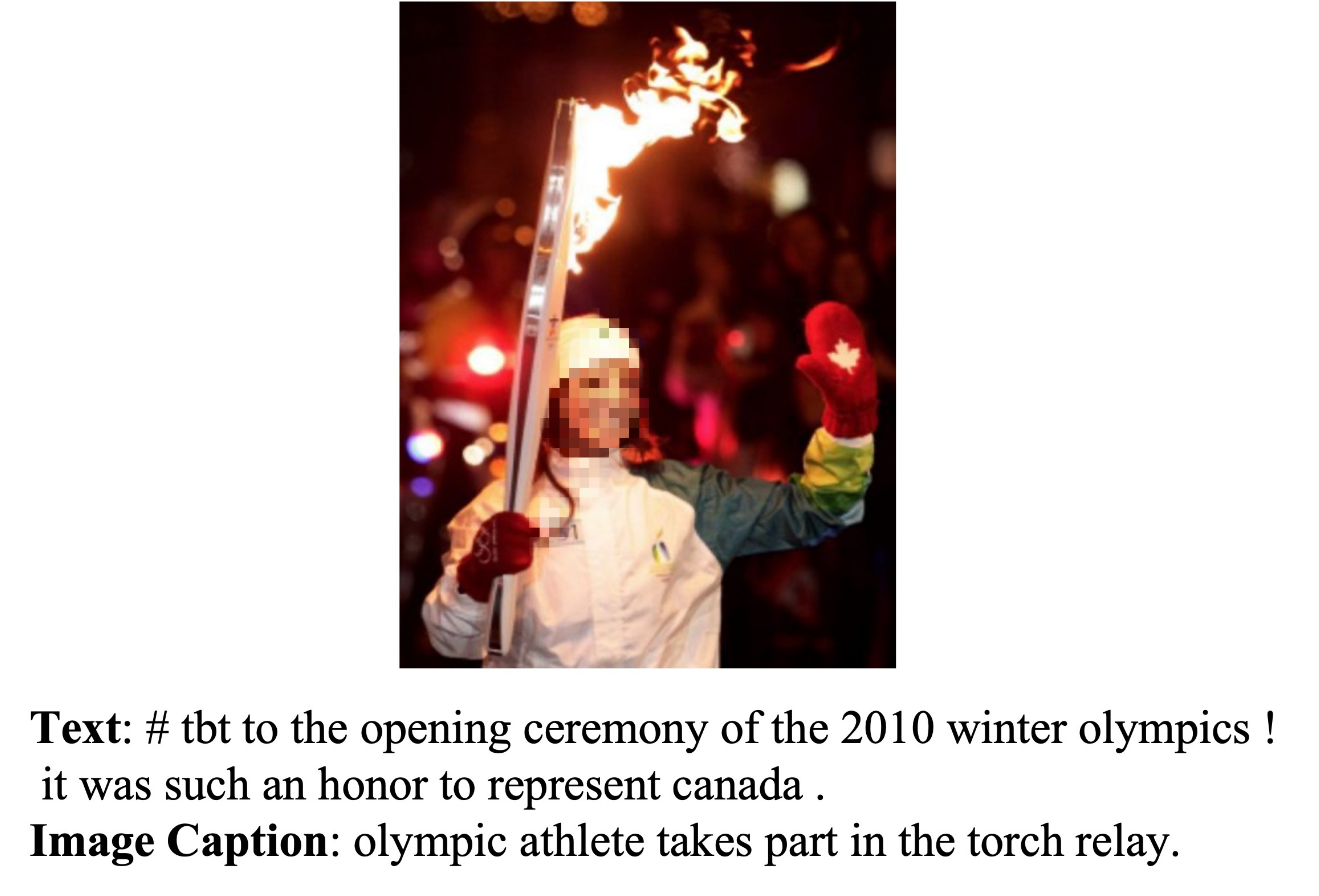} 
\centering
\caption{Non-Sarcasm}
\label{casestudy2}
\end{subfigure}
\caption{Wrongly detected samples by our framework without external knowledge.}\label{casestudy}
\vspace{-10pt}
\end{figure}

To further justify the effectiveness of external knowledge for sarcasm detection task, we provide case studies on the samples that are incorrectly predicted by our proposed framework with only text-image pair but can be accurately classified with the help of image captions. 
For example, intravenous injection depicted in Figure \ref{casestudy1} can indicate a flu or hospitalization via image modeling, which aligns with \textit{flu} or \textit{hospital} in the text input. However, by generating an image caption expressing a bad mood indicated by \textit{suffered}, it becomes easy to detect the sarcastic nature of this sample by contrasting \textit{fun} in the text description and \textit{suffer} in the image caption. As another example shown in Figure \ref{casestudy2}, the image encoder only detects a human holding a torch without any contexts and wrongly predicts the sample as sarcasm because of the disalignment between the image and text description. By generating the image caption expressing an \textit{olympic athlete}, the knowledge-fused model is able to detect the alignment and correctly classifies this sample. This reflects that by further utilizing CLIP \cite{CLIP} and GPT-2 \cite{GPT-2} models pre-trained using large-scale data as an external knowledge source, the generated image captions are more expressive to understand some sophisticated visual concepts and to mitigate the furtiveness and subtlety of sarcasm.
\begin{figure}[t]
\begin{subfigure}[b]{0.15\textwidth}
\centering
\includegraphics[width=\linewidth]{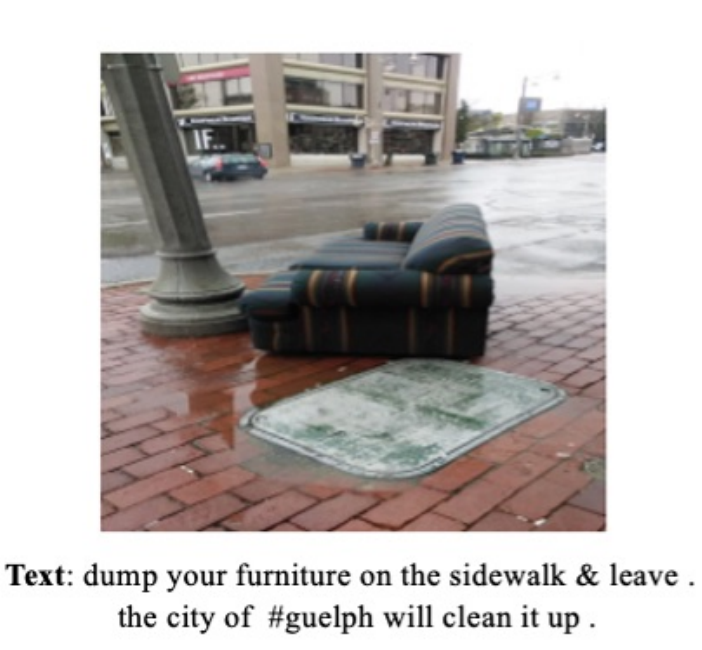}
\caption{Sarcasm}
\label{sarcasmcase}
\end{subfigure}
\begin{subfigure}[b]{0.32\textwidth}
\includegraphics[width=\linewidth]{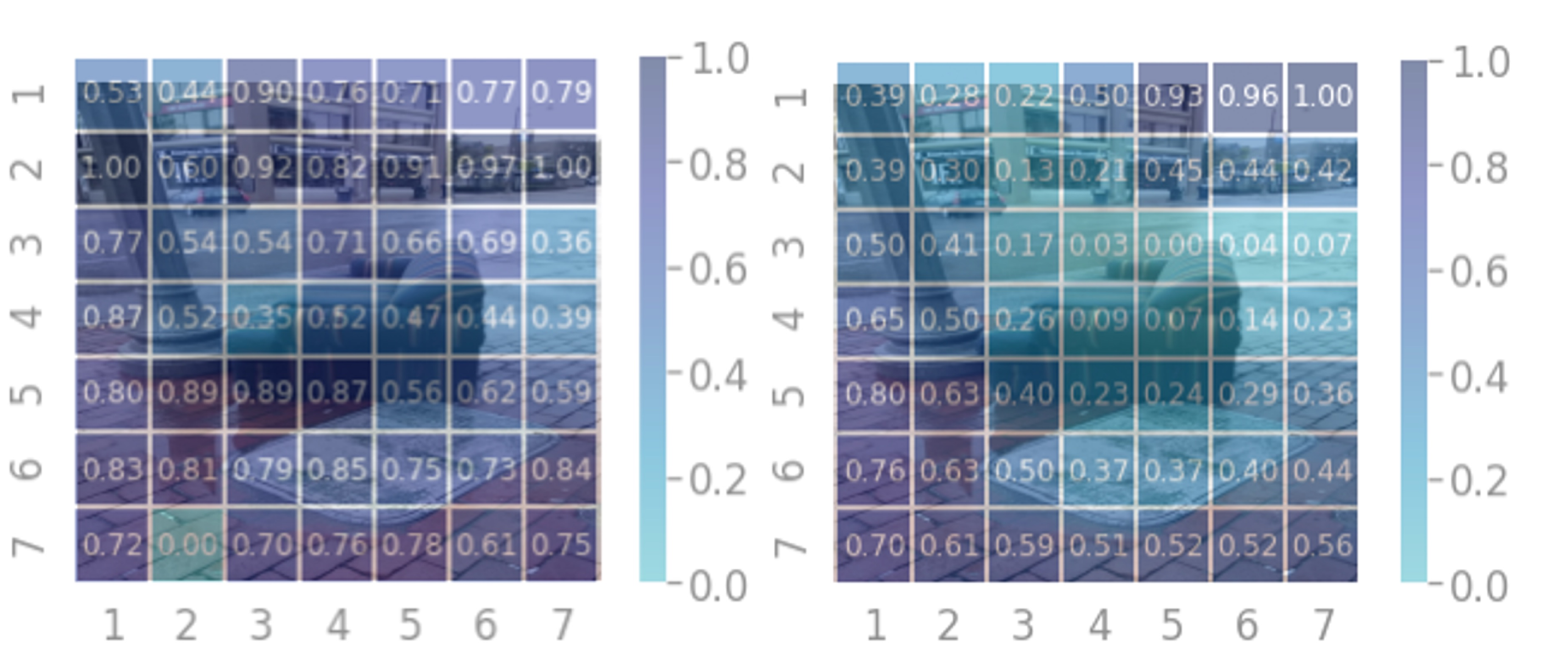} 
\centering
\caption{Congruity score visualization}
\label{textimage}
\end{subfigure}
\caption{Visualization of atomic-level and composition-level congruity score between text and image. Better to be viewed in color format and zoom.}\label{vis}
\vspace{-15pt}
\end{figure}

We further illustrate the effectiveness of our hierarchical modeling by showing the congruity score maps in Figure \ref{vis}. Given a sarcastic sample in Figure \ref{sarcasmcase}, we
visualize the congruity scores between the text and image in both atomic-level module $\mathbf{s}_a$ (left side of Figure \ref{textimage}) and composition-level module $\mathbf{s}_p$ (right side of Figure \ref{textimage}). The smaller the values, the less similar between the text and image (i.e., more likely to be detected as sarcasm). It can be shown that the atomic-level module  attends to \textit{furniture} in the image whereas the composition-level module down-weighs those patches, making the text and image less similar for sarcasm prediction. Correspondingly, our proposed hierarchical structure has the power to refine atomic congruity to identify more complex mismatches for multi-modal sarcasm detection using graph neural networks.
\section{Conclusion}
In this paper, we propose to tackle the problem of sarcasm detection by reasoning atomic-level congruity and composition-level congruity in a hierarchical manner. Specifically, we propose to model the atomic-level congruity based on the multi-head cross attention mechanism and the composition-level congruity based on graph attention networks. In addition, we propose to exploit the effect of various knowledge resources on enhancing the discriminative power of the model. Evaluation results demonstrate the superiority of our proposed model and the benefit of image captions as external knowledge for sarcasm detection.

\section*{Limitations}
We present two possible limitations: 1) we only use the Twitter dataset for evaluation. However, to the best of our knowledge, this dataset is the only benchmark for the evaluation of multi-modal sarcasm detection in our community. Nevertheless, we conduct extensive experiments with various metrics to show the superiority of our proposed method. We leave the construction of more high-quality benchmarks in our future work; 2) our knowledge enhancement strategy in Section \ref{knowledge_section} may not be suitable for ANPs and Image Attributes.  We analyze the results in Section \ref{common_know}.  Consequently, there is a pressing need for a more general and elegant knowledge integration method in view of the importance of external knowledge for multi-modality sarcasm detection.


\section*{Ethics Statement}
This paper is informed by the ACM Code of Ethics and Professional Conduct. Firstly, we respect valuable and creative works in sarcasm detection and other related research domains. We especially cite relevant papers and sources of pre-trained models and toolkits exploited by this work as detailed and reasonable as possible. Besides, we will release our code based on the licenses of any used artifacts. Secondly, our adopted dataset does not include sensitive privacy individual information and will not introduce any information disorder to society. For precautions to prevent re-identification of data, we mask facial information in Figure \ref{casestudy2}. At last, as our proposed sarcasm detection method benefits the identification of authentic intentions in multi-modal posts on social media, we expect our proposed method can also bring positive impact on related problems, such as opinion mining, recommendation system, and information forensics in the future.

\section*{ACKNOWLEDGEMENT}
This work was supported in part by CityU New Research Initiatives/Infrastructure Support from Central (APRC 9610528), the Research Grant Council (RGC) of Hong Kong through Early Career Scheme (ECS) under the Grant 21200522 and Hong Kong Innovation and Technology Commission (InnoHK Project CIMDA).

\bibliographystyle{acl_natbib}
\bibliography{anthology,ref}

\appendix
\begin{figure*}[h]
  \centering
  \includegraphics[width=\linewidth]{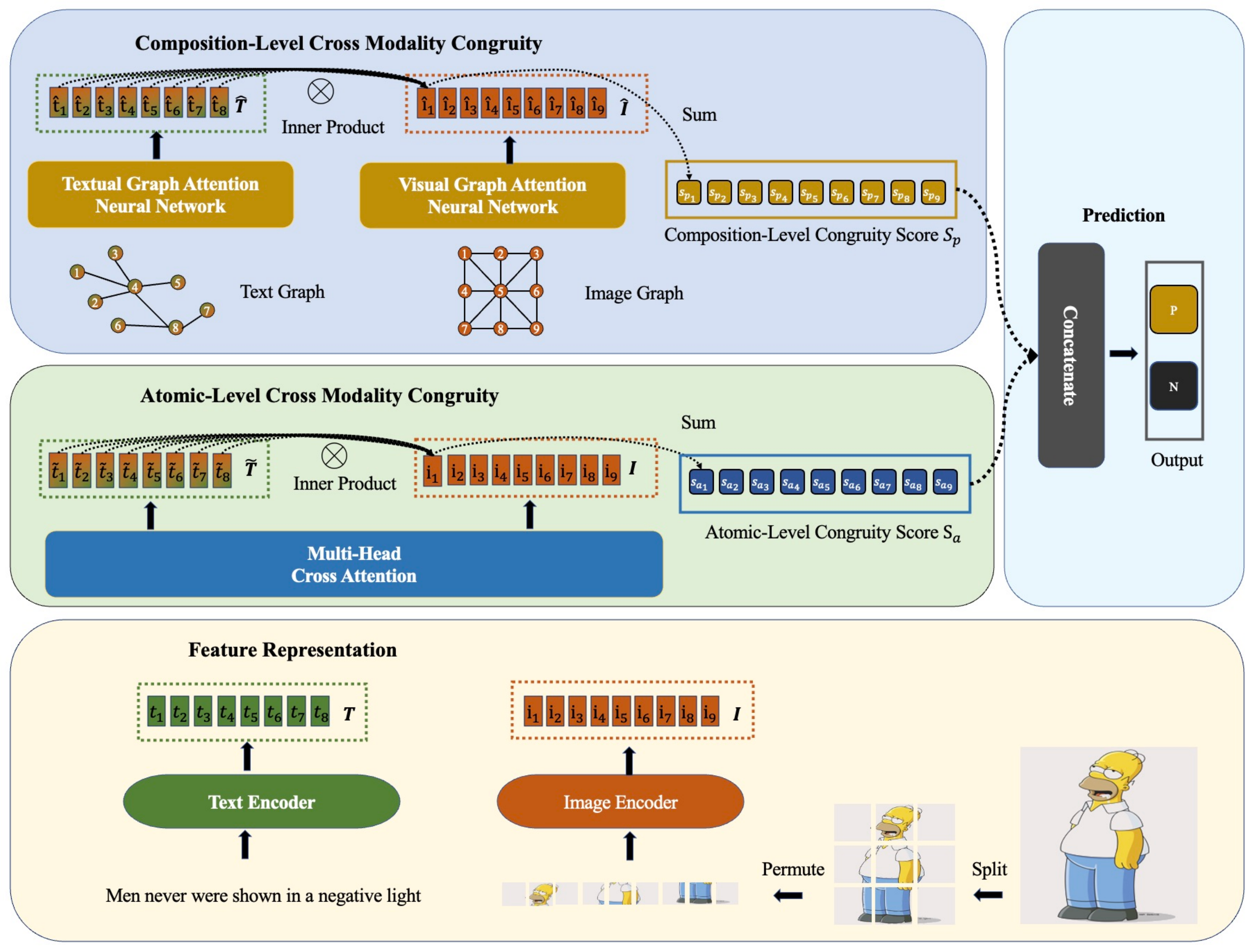}
  \caption{An overview of text-image branch of our approach, which involves three main modules: Feature Representation, Atomic-Level Cross Modality Congruity and Composition-Level Cross Modality Congruity.}
\label{overview}
\end{figure*}
\section{Model Overview}
For illustration, we give a figure of the text-image branch that can capture atomic-level and composition-level congruity between textual and visual modalities for multimodal sarcasm detection. Specifically, this figure is comprised of three main components: Feature Representation, Atomic-Level Cross Modality Congruity and  Composition-Level Cross Modality Congruity, where Feature Representation extracts feature representations corresponding to texts and images, Atomic-Level Cross Modality Congruity obtains congruity scores via a
multi-head cross-attention mechanism, and Composition-Level Cross Modality Congruity produces composition-level congruity scores based on constructed textual and visual graphs.
\section{Modal Parameters}
\label{Modal Parameters}
\begin{table}[h]
 \caption{Parameters of our model}
 \label{model parameters}
  \begin{adjustbox}{max width=0.8\columnwidth}
  \begin{tabular}{c|cc}
    \hline
    Parameters&Value\\
    \hline
    Max length of text&$100$\\
    Max length of image caption&$20$\\
    MCA layers for text-image branch&$6$\\
    MCA layers for text-knowledge branch&$3$\\
    Head number of MCA&$5$\\
    GAT layers for text-image branch&$2$\\
    GAT layers for text-knowledge branch&$2$\\
    Batch size&32\\
    Learning rate& $2e\!\!-\!\!5$\\
    Weight decay &$5e\!\!-\!\!3$\\
    Dropout rate & $0.5$\\
    \hline
\end{tabular}
\end{adjustbox}
\end{table}

For our model, the max length of sarcasm text is set to $100$ and the max length of generated image caption is set to $20$. For the architecture, the number of the multihead cross-attention layer is set to $6$ for text-image branch and $3$ for text-knowledge branch to capture atomic-level congruity score. The head number is set to $5$. The number of graph attention layer is set to $2$ to obtain composition-level congruity score for both branches. We use Adam optimizer with a learning rate of $2e\!\!-\!\!5$, weight decay of $5e\!\!-\!\!3$, batch size as 32 and dropout rate as 0.5 to train the model. For more detail, please refer to the checklist in our implementation.

\end{document}